\documentclass[11pt,a4paper]{article}
\usepackage[hyperref]{emnlp2018}
\usepackage{times}
\usepackage{latexsym}

\usepackage{amssymb}% http://ctan.org/pkg/amssymb
\usepackage{pifont}

\usepackage{url}

\usepackage{times}
\usepackage{url}
\usepackage{latexsym}
\usepackage{amssymb}
\usepackage{amsmath}
\DeclareMathOperator*{\argmax}{argmax} % thin space, limits underneath in displays

\usepackage{cleveref}

\usepackage{stmaryrd}
\usepackage{graphicx}
\usepackage{adjustbox}
\usepackage{listings}
\usepackage{caption,setspace}
\usepackage{diagbox} % package for diagonal box in table 
\usepackage{booktabs} % package for toprule, midrule, bottomrule
\usepackage{rotating}
\usepackage{array,multirow}
\usepackage{hhline}
\usepackage{todonotes}
\usepackage{flushend}
\usepackage[inline]{enumitem}
\usepackage{csquotes}
\usepackage{subcaption}

\usepackage{color}
\newcommand{\cmark}{\ding{51}}%
\newcommand{\xmark}{\ding{55}}%

\newcommand{\eg}{e.g., }
\newcommand{\ie}{i.e., }
\newcommand{\figref}[1]{Fig.~\ref{#1}}    % within sentence
\newcommand{\Figref}[1]{Figure~\ref{#1}}  % start of sentence
\newcommand{\tabref}[1]{Table~\ref{#1}}

\newcommand{\secref}[1]{Section~\ref{#1}}
\newcommand{\equref}[1]{Eq.~(\ref{#1})}

\usepackage{mathtools}
\newcommand{\prob}{\operatorname{\mathbb{P}}\probarg}
\DeclarePairedDelimiterX{\probarg}[1]{(}{)}{%
  \ifnum\currentgrouptype=16 \else\begingroup\fi
  \activatebar#1
  \ifnum\currentgrouptype=16 \else\endgroup\fi
}

\newcommand{\innermid}{\nonscript\;\delimsize\vert\nonscript\;}
\newcommand{\activatebar}{%
  \begingroup\lccode`\~=`\|
  \lowercase{\endgroup\let~}\innermid 
  \mathcode`|=\string"8000
}

\newcommand{\norm}[1]{\left\lVert#1\right\rVert}

\usepackage{siunitx}
\aclfinalcopy % Uncomment this line for the final submission

\title{Adversarial training for multi-context\\ 
joint entity and relation extraction}

\author{Giannis Bekoulis\qquad Johannes Deleu\qquad Thomas Demeester\qquad Chris Develder\\Ghent University -– imec, IDLab\\Department of Information Technology\\ \tt{firstname.lastname@ugent.be}}

\date{}

\begin{document}
\maketitle
\begin{abstract}
Adversarial training (AT) is a regularization method that can be used to improve the robustness of neural network methods by adding small perturbations in the training data. We show how to use AT for the tasks of entity recognition and relation extraction. In particular, we demonstrate that applying AT to a general purpose baseline model for jointly extracting entities and relations, allows improving the state-of-the-art effectiveness on several datasets in different contexts (\ie news, biomedical, and real estate data) and for different languages (English and Dutch).

\end{abstract}

\section{Introduction}
Many neural network methods have recently been exploited in various natural language processing (NLP) tasks, such as parsing~\cite{zhang:16}, POS tagging~\cite{lample:16}, relation extraction~\cite{santos:15}, translation~\cite{bahdanau:14}, and joint tasks~\cite{miwa:16}. 
However, \newcite{szegedy:13} observed that intentional small scale perturbations (\ie adversarial examples) to the input of such models may lead to incorrect decisions (with high confidence). 
\newcite{goodfellow:15} proposed adversarial training (AT) (for image recognition) as a regularization method which uses a mixture of clean and adversarial examples to enhance the robustness of the model. Although AT has recently been applied in NLP tasks %(see \secref{sec:related_work})
(e.g., text classification~\cite{miyato:17}), this paper --- to the best of our knowledge --- is the first attempt  investigating regularization effects of AT in a joint setting for two related tasks.

We start from a baseline joint model that performs the tasks of named entity recognition (NER) and relation extraction at once. 
Previously proposed models~(summarized in \secref{sec:related_work}) exhibit several issues that the neural network-based baseline approach (detailed in \secref{sec:model}) overcomes:
\begin{enumerate*}[label=(\roman*)]
\item our model uses automatically extracted features without the need of external parsers nor manually extracted features~(see \newcite{gupta:16, miwa:16, li:17}),
\item all entities and the corresponding relations within the sentence are extracted at once, instead of examining one pair of entities at a time (see~\newcite{heike:17}), and
\item we model relation extraction in a multi-label setting, allowing multiple relations per entity (see \newcite{katiyar:17, bekoulis:18}).
\end{enumerate*}
The core contribution of the paper is the use of AT
% Next, we explain how AT can be used
as an extension in the training procedure for the joint extraction task (\secref{sec:adversarial}).

To evaluate the proposed AT method, we perform a large scale experimental study in this joint task (see \secref{sec:setup}), using datasets from different contexts (\ie news, biomedical, real estate) and languages (\ie English, Dutch). 
We use a strong baseline that outperforms all previous models that rely on automatically extracted features, achieving state-of-the-art performance (\secref{sec:results}).
Compared to the baseline model, applying AT during training leads to a consistent additional increase in joint extraction effectiveness.

\section{Related work}

\label{sec:related_work}

\textbf{Joint entity and relation extraction:} Joint models~\citep{li:14,miwa:14} that are based on manually extracted features have been proposed for performing both the named entity recognition (NER) and relation extraction subtasks at once. These methods rely on the availability of NLP tools (\eg POS taggers) or manually designed features leading to additional complexity. Neural network methods have been exploited to overcome this feature design issue and usually involve RNNs and CNNs~\citep{miwa:16,zheng:17}. 
Specifically,~\newcite{miwa:16} as well as \newcite{li:17} apply
bidirectional tree-structured RNNs for different contexts (\ie news, biomedical) to capture syntactic information (using external dependency parsers). 
\newcite{gupta:16} propose the use of various manually extracted features along with RNNs. \newcite{heike:17} solve the simpler problem of entity classification (EC, assuming entity boundaries are given), instead of NER, and they replicate the context around the entities, feeding entity pairs to the relation extraction layer.
\newcite{katiyar:17} investigate RNNs with attention without taking into account that relation labels are not mutually exclusive.
Finally,~\newcite{bekoulis:18} use LSTMs in a joint model for extracting just one relation at a time, but increase the complexity of the NER part.
Our baseline model enables simultaneous extraction of multiple relations from the same input. Then, we further extend this strong baseline using adversarial training.

\textbf{Adversarial training (AT)} \cite{goodfellow:15} has been proposed to make classifiers more robust to input perturbations in the context of image recognition. 
In the context of NLP, several variants have been proposed for different tasks such as text classification~\cite{miyato:17}, relation extraction~\cite{wu:17} and POS tagging~\cite{yasunaga:17}. 
AT is considered as a regularization method. 
Unlike other regularization methods (\ie dropout~\cite{srivastava:14}, word dropout~\cite{iyyer:15}) that introduce random noise, AT generates perturbations that are variations of examples easily misclassified by the model.
\begin{figure}%[!ht]
\includegraphics[width=\columnwidth]{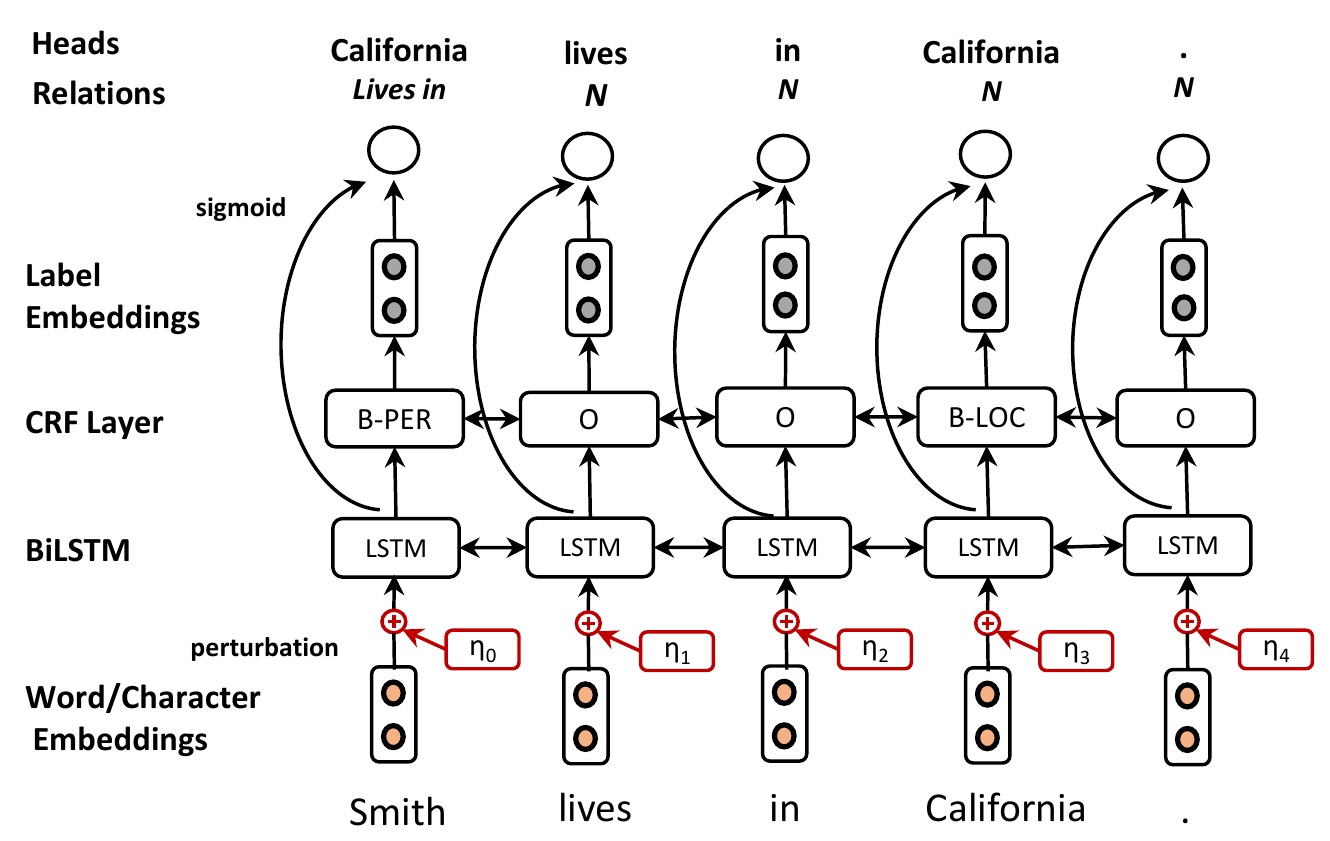}
\caption{Our model for joint entity and relation extraction with adversarial training (AT) comprises
\begin{enumerate*}[label=(\roman*)]
\item a word and character embedding layer,
\item a BiLSTM layer,
\item a CRF layer and 
\item a relation extraction layer.
\end{enumerate*}
In AT, we compute the worst-case perturbations $\eta$ of the input embeddings.
}

\label{fig:model}
\end{figure}

\section{Model}

\subsection{Joint learning as head selection}
\label{sec:model}
\noindent
The baseline model, described in detail in \newcite{bekoulis:18b}, is illustrated in \figref{fig:model}. It
aims to detect
\begin{enumerate*}[label=(\roman*)]
\item the type and the boundaries of the entities and
\item the relations between them.
\end{enumerate*}
The input is a sequence of tokens (\ie sentence) $w={w_1,...,w_n }$. 
We use character level embeddings to implicitly capture morphological features (\eg prefixes and suffixes), representing each character by a vector (embedding). The character embeddings are fed to a bidirectional LSTM (BiLSTM) to 
obtain the character-based representation of the word.
We also use pre-trained word embeddings. Word and character embeddings are concatenated to form the final token representation, which is then fed to a BiLSTM layer to extract sequential information.

For the \textbf{NER task}, we adopt the BIO (Beginning, Inside, Outside) encoding scheme. 
In \figref{fig:model}, the B-\textit{PER} tag is assigned to the beginning token of a `person' (\emph{PER}) entity. For the prediction of the entity tags, we use:
\begin{enumerate*}[label=(\roman*)]
\item a softmax approach for the entity classification (EC) task (assuming entity boundaries given) or
\item a CRF approach where we identify both the \emph{type} and the boundaries for each entity.
\end{enumerate*}
During decoding, in the softmax setting, we greedily detect the entity \emph{types} of the tokens (\ie independent prediction). Although  independent distribution of \emph{types} is reasonable for EC tasks, this is not the case when there are strong correlations between neighboring tags. For instance, the BIO encoding scheme imposes several constraints in the NER task (\eg the B-\textit{PER} and I-\textit{LOC} tags cannot be sequential). Motivated by this intuition, we use a linear-chain CRF for the NER task~\cite{lample:16}.
For decoding, in the CRF setting, we use the Viterbi algorithm. 
During training, for both EC (softmax) and NER tasks (CRF), we minimize the cross-entropy loss $\mathcal{L}_{\textsc{ner}}$.
The entity tags are later fed into the relation extraction layer as label embeddings (see \figref{fig:model}), assuming that knowledge of the entity \emph{types} is beneficial in predicting the relations between the involved entities. 

We model the \textbf{relation extraction task} as a multi-label head selection problem~\citep{bekoulis:18b,zhang:16}. In our model, each word $w_i$ can be involved in multiple relations with other words. For instance, in the example illustrated in \figref{fig:model}, ``Smith'' could be involved not only in a \emph{Lives in} relation with the token ``California'' (head) but also in other relations simultaneously (\eg \emph{Works for, Born In} with some corresponding tokens).
The goal of the task is to predict for each word $w_i$, a vector of heads $\hat{y}_i$ and the vector of corresponding relations $\hat{r}_i$.
We compute the score $s(w_j , w_i, r_k)$ of word $w_j$ to be the head of $w_i$ given a relation label $r_k$ using a single layer neural network.
The corresponding probability is defined as:
$\prob*{w_j,r_k |w_i;\theta}=\sigma(s(w_j , w_i,r_k))$, where $\sigma(.)$ is the sigmoid function. 
During training, we minimize the cross-entropy loss $\mathcal{L}_{\textrm{rel}}$ as:
\begin{equation}
\sum_{i=0}^{n}\sum_{j=0}^{m}{-\log \prob*{y_{i,j},\,r_{i,j}|w_i;\theta}}
\end{equation}
where $m$ is the number of associated heads (and thus relations) per word $w_i$. During decoding, the most probable heads and relations are selected using threshold-based prediction. 
The final objective for the joint task is computed as 
$\mathcal{L}_\textsc{Joint}(w;\theta)=\mathcal{L}_\textsc{ner} + \mathcal{L}_\textrm{rel}$ where $\theta$ is a set of parameters. In the case of multi-token entities, only the last token of the entity can serve as head of another token, to eliminate redundant relations. If an entity is not involved in any relation, we predict the auxiliary ``N'' relation label and the token itself as head.

\subsection{Adversarial training (AT)}
\label{sec:adversarial}

We exploit the idea of AT~\cite{goodfellow:15} as a regularization method to make our model robust to input perturbations. Specifically, we generate examples which are variations of the original ones by adding some noise at the level of the concatenated word representation~\cite{miyato:17}. This is similar to the concept introduced by~\newcite{goodfellow:15} to improve the robustness of image recognition classifiers.
We generate an adversarial example by adding the worst-case
perturbation $\eta_{adv}$ to the original embedding $w$ that maximizes the loss function:
\begin{equation}
\eta_{adv}=\argmax_{\norm{\eta}\leq \epsilon }\mathcal{L}_\textsc{Joint}(w+\eta;\hat{\theta})
\label{eq:eta_adv}
\end{equation}
where $\hat{\theta}$ is a copy of the current model parameters. Since \equref{eq:eta_adv} is intractable in neural networks, we use
the approximation proposed in~\newcite{goodfellow:15} defined as:
$\eta_{adv}=\epsilon g/\norm{g}\text{, with } g=\nabla_{w}\mathcal{L}_\textsc{Joint}(w;\hat{\theta})$, where $\epsilon$ is a small bounded norm treated as a hyperparameter. 
Similar to~\newcite{yasunaga:17}, we set $\epsilon$ to be $\alpha\sqrt{D}$
(where $D$ is the dimension of the embeddings). 
We train on the mixture of original
and adversarial examples, so the final loss is computed as:
$ \mathcal{L}_\textsc{Joint}(w;\hat{\theta}) +\mathcal{L}_\textsc{Joint}(w+\eta_{adv};\hat{\theta})$.

\section{Experimental setup}
\label{sec:setup}
\noindent We evaluate our models on four datasets, using the code as available from
our github codebase.\footnote{\url{https://github.com/bekou/multihead_joint_entity_relation_extraction}}
Specifically, we follow the 5-fold cross-validation defined by~\newcite{miwa:16} for the ACE04~\citep{doddington:04} dataset. 
For the CoNLL04~\citep{roth:04} EC task (assuming boundaries are given), we use the same splits as in~\newcite{gupta:16,heike:17}. We also evaluate our models on the NER task similar to \newcite{miwa:14} in the same dataset using 10-fold cross validation. 
For the Dutch Real Estate Classifieds, DREC~\citep{bekoulis:17} dataset, we use train-test splits as in \newcite{bekoulis:18}.
For the Adverse Drug Events, ADE~\citep{gurulingappa:12b}, we perform 10-fold cross-validation similar to \newcite{li:17}. To obtain comparable results that are not affected by the input embeddings, we use the embeddings of the previous works. 
We employ early stopping in all of the experiments. We use the Adam optimizer~\citep{kingma:14} and we fix the hyperparameters (\ie $\alpha$, dropout values, best epoch, learning rate) on the validation sets. 
The scaling parameter $\alpha$ is selected from \{\num{5e-2}, \num{1e-2}, \num{1e-3}, \num{1e-4}\}. 
Larger values of $\alpha$ (\ie larger perturbations) lead to consistent performance decrease in our early experiments. This can be explained from the fact that adding more noise can change the content of the sentence as also reported by~\newcite{wu:17}.

We use three types of evaluation, namely: 
\begin{enumerate*}[label=(\roman*)]
\item \emph{S(trict)}: we score an entity as correct if both the entity boundaries and the entity \emph{type} are correct (ACE04, ADE, CoNLL04, DREC),
\item \emph{B(oundaries)}: we score an entity as correct if only the entity boundaries are correct while the entity \emph{type} is not taken into account (DREC) and
\item \emph{R(elaxed)}: a multi-token entity is considered correct if
at least one correct \emph{type} is assigned to the tokens comprising the entity, assuming that the boundaries are known (CoNLL04),
\end{enumerate*}
to compare to previous works. In all cases, a relation is considered as correct when both the relation \emph{type} and the argument entities are correct.

\begin{table}%[!ht]
\centering
\resizebox{\columnwidth}{!}{%

\begin{tabular}{@{\extracolsep{4pt}}ccccccc@{}} % trick for spacing between clines
 \toprule
 %& \multicolumn{1}{c}{} & \multicolumn{1}{c}{}  & \multicolumn{1}{c}{} &  \multicolumn{3}{c}{F$_1$}  \\
%\cline{5-7}

%\cline{7-9}

 & \multicolumn{1}{c}{Settings}& \multicolumn{1}{c}{Features}& \multicolumn{1}{c}{Eval.} & \multicolumn{1}{c}{Entity} & \multicolumn{1}{c}{Relation}& \multicolumn{1}{c}{Overall}  \\

 \midrule
\parbox[c]{5mm}{\multirow{4}{*}{\rotatebox[origin=c]{90}{\parbox{1.4cm}{\centering ACE\\04}}}}

&\newcite{miwa:16}&\cmark &   \emph{S}    & 81.80    & 48.40 & 65.10  \\
&\newcite{katiyar:17} &\xmark&   \emph{S}    &79.60   & 45.70 & 62.65\\

&  baseline&\xmark & \emph{S}   &81.16 & 47.14 & 64.15\\
&  baseline + AT&\xmark & \emph{S}  &\textbf{81.64} & \textbf{47.45} & \textbf{64.54}
\\

 \midrule
\parbox[c]{5mm}{\multirow{8}{*}{\rotatebox[origin=c]{90}{\parbox{1.0cm}{\centering CoNLL\\04}}}}

 &\newcite{gupta:16}&\cmark &   \emph{R}   &92.40  &69.90 &81.15 \\
&\newcite{gupta:16}&\xmark &   \emph{R}    &88.80  &58.30 &73.60  \\
&\newcite{heike:17}&\xmark &   \emph{R}    &82.10  &62.50 &72.30    \\
&  baseline EC&\xmark & \emph{R}        &\textbf{93.26}  &67.01 & 80.14\\
&  baseline EC + AT&\xmark & \emph{R}  & 93.04    &\textbf{67.99} & \textbf{80.51}\\
% \cline{2-11}
%&\cite{miwa:14}&\cmark &   \emph{strict} &81.20 & 80.20 & 80.70 &76.00 & 50.90 & 61.00 & 70.85  \\
%&baseline&\xmark &   \emph{strict}  &83.75  &84.06       &83.90     & 63.75&60.43&62.04&72.97\\
%&baseline + AT&\xmark &   \emph{strict}  &84.86  &85.17       &\textbf{85.01}     & 70.36&55.69&\textbf{62.17}&\textbf{73.59}\\
 \cline{2-7}
&\newcite{miwa:14}&\cmark    & \emph{S}    & 80.70     & 61.00                 & 70.85  \\
&baseline&\xmark &   \emph{S}          & 83.04      &61.04                  &72.04\\
&baseline + AT&\xmark &   \emph{S}    & \textbf{83.61}   & \textbf{61.95} & \textbf{72.78}\\
\midrule
\parbox[c]{5mm}{\multirow{5}{*}{\rotatebox[origin=c]{90}{\parbox{1.0cm}{\centering DREC}}}}
&\newcite{bekoulis:18} &\xmark&   \emph{B}  &79.11 &49.70 &64.41 \\
&  baseline &\xmark& \emph{B}     &82.30   &52.81& 67.56\\
&  baseline + AT &\xmark& \emph{B}       &\textbf{82.96}    &\textbf{53.87}& \textbf{68.42}\\

 \cline{2-7}
&baseline &\xmark&  \emph{S}      &81.39    &52.26& 66.83  \\
&baseline + AT &\xmark&  \emph{S}       &\textbf{82.04}   &\textbf{53.12} & \textbf{67.58} \\

\midrule
\parbox[c]{5mm}{\multirow{4}{*}{\rotatebox[origin=c]{90}{\parbox{1.4cm}{\centering ADE}}}}

&\newcite{li:16}&\cmark &   \emph{S}       &79.50     & 63.40 & 71.45\\
&\newcite{li:17}&\cmark &   \emph{S}          &84.60     & 71.40 & 78.00 \\

&  baseline&\xmark & \emph{S}  &86.40    &74.58 & 80.49\\
&  baseline + AT&\xmark & \emph{S}    &\textbf{86.73}  &\textbf{75.52} & \textbf{81.13}\\

%\hhline{~=======}

%\hhline{~=======}

%\hhline{~=======}
\bottomrule
\end{tabular}
 }

 \caption{Comparison of our method with the state-of-the-art in terms of F$_1$ score.
The proposed models are:
 (i)~baseline,
(ii)~baseline EC (predicts only entity classes) and
(iii)~baseline (EC) + AT (regularized by AT). The \cmark and \xmark~symbols indicate whether the models rely on external NLP tools. We include different evaluation types (\emph{S}, \emph{R} and \emph{B}).}
%\vspace{-0.1em}
\label{tab:results}
 \end{table}

\begin{figure*}
  
\centering
        \begin{subfigure}{0.475\textwidth}
            \centering
             \includegraphics[trim={0cm 0cm 0 0},clip,width=\linewidth]{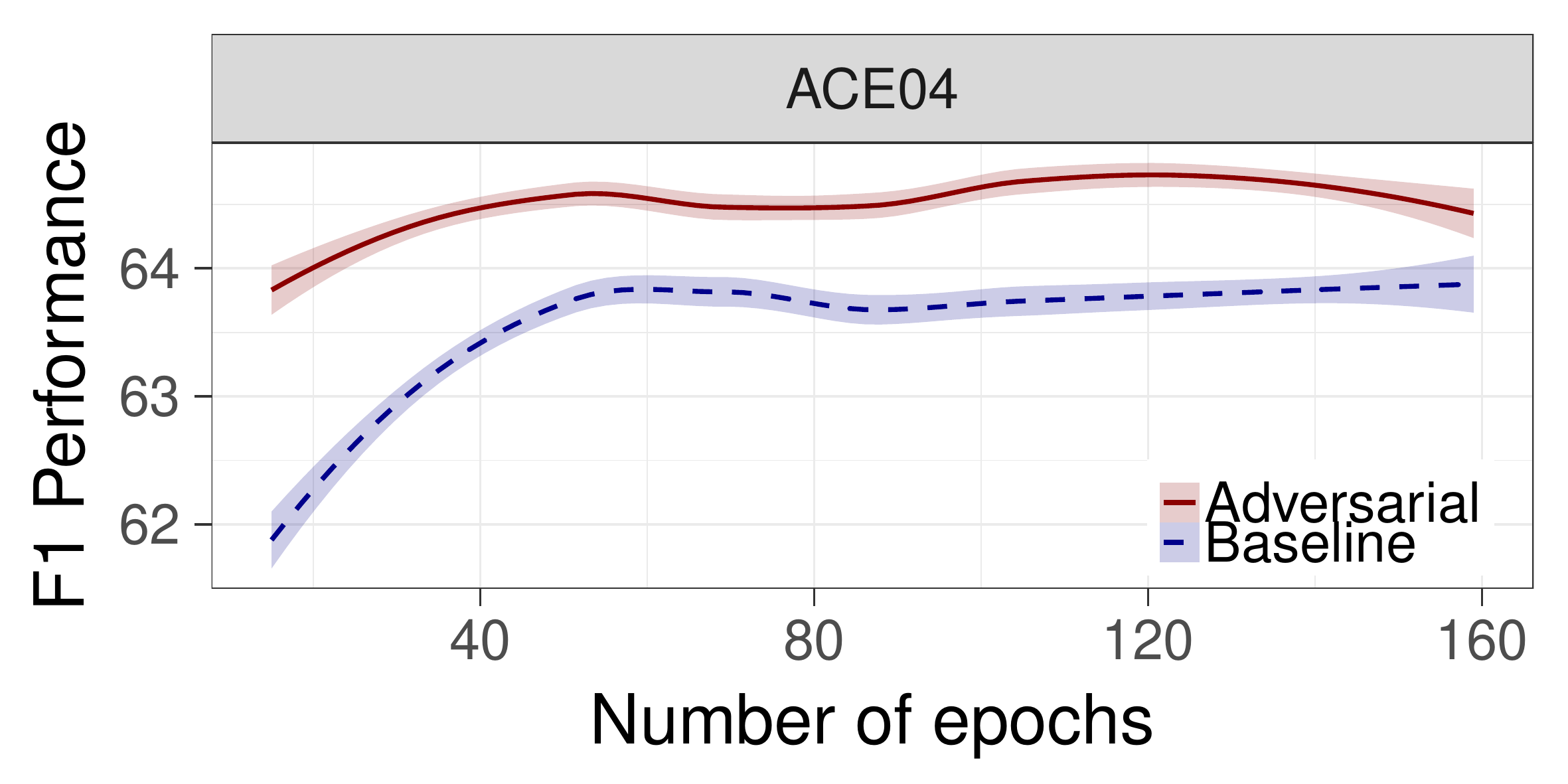}
  %\caption{ACE04}
            %\label{fig:sub1}
        \end{subfigure}
        \hfill
        \begin{subfigure}{0.465\textwidth}  
            \centering 
            \includegraphics[trim={0cm 0cm 0 0},clip,width=\linewidth]{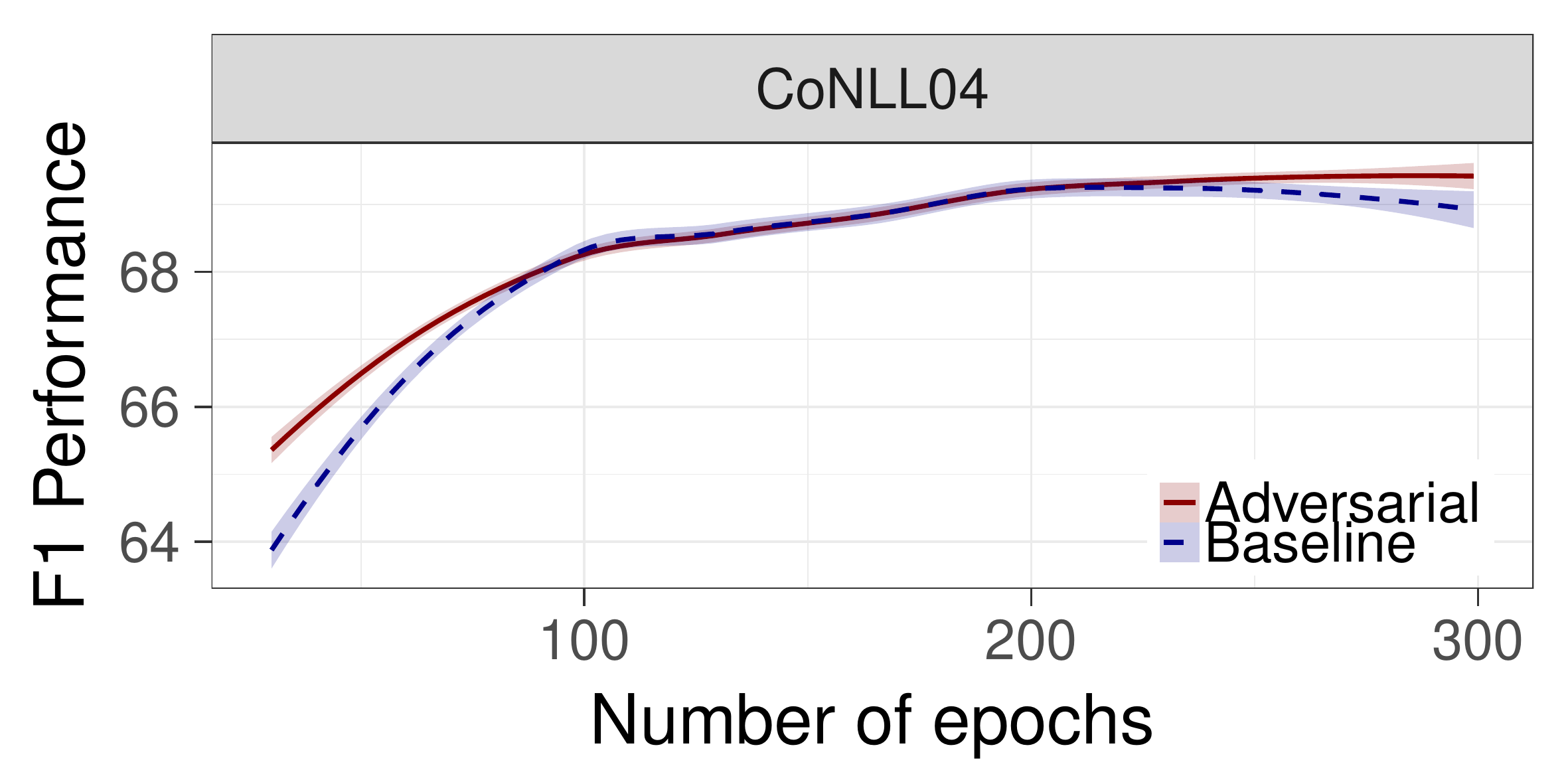}
  %\caption{CoNLL04}
  %\label{fig:sub2}
        \end{subfigure}
                \begin{subfigure}{0.475\textwidth}   
            \centering 
             \includegraphics[trim={0 0 0 0},clip,width=\linewidth]{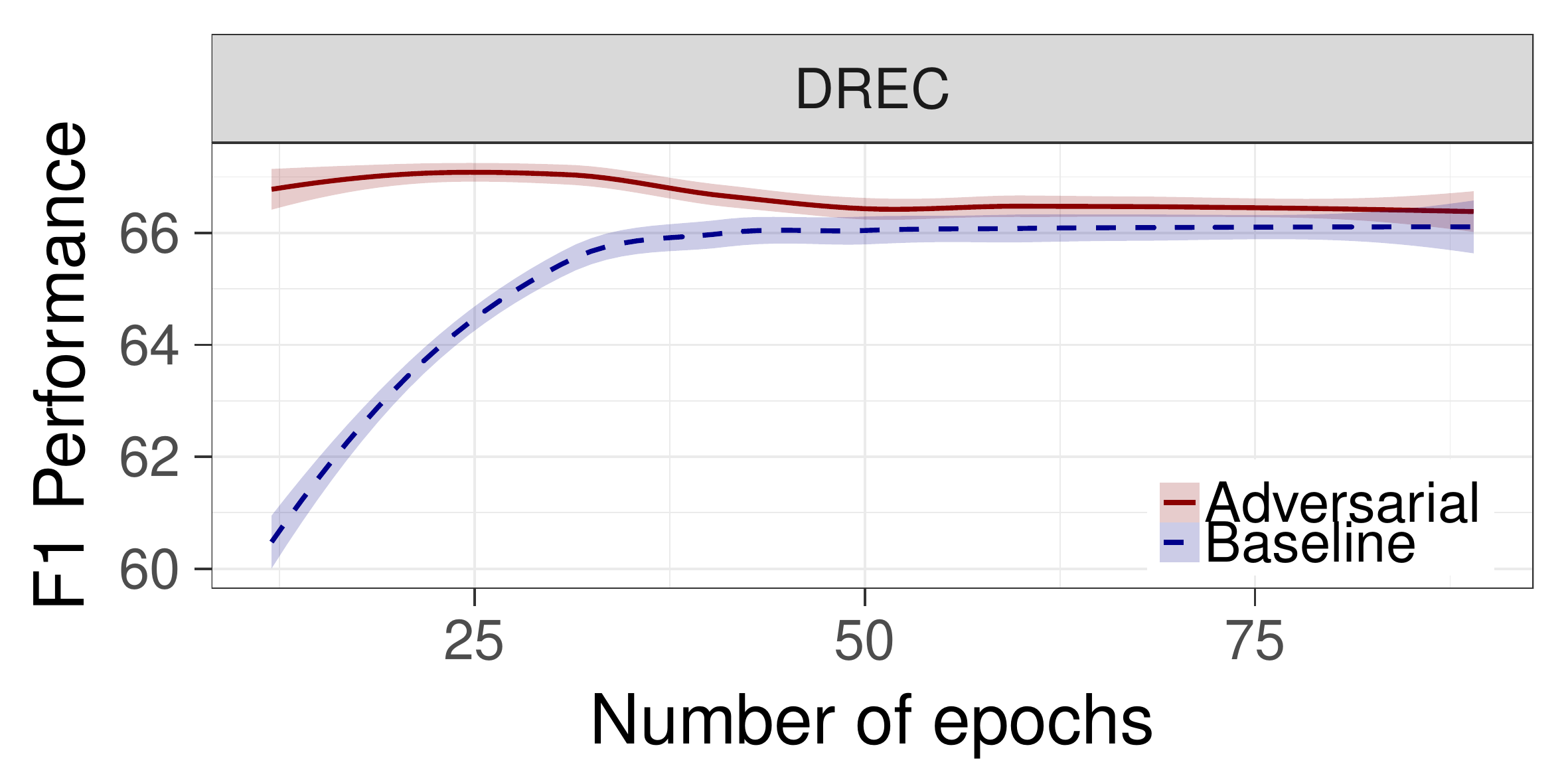}
  %\caption{DREC}
  %\label{fig:sub3}
        \end{subfigure}
        \hfill
        \begin{subfigure}{0.475\textwidth}   
            \centering 
            \includegraphics[trim={0cm 0 0 0},clip,width=\linewidth]{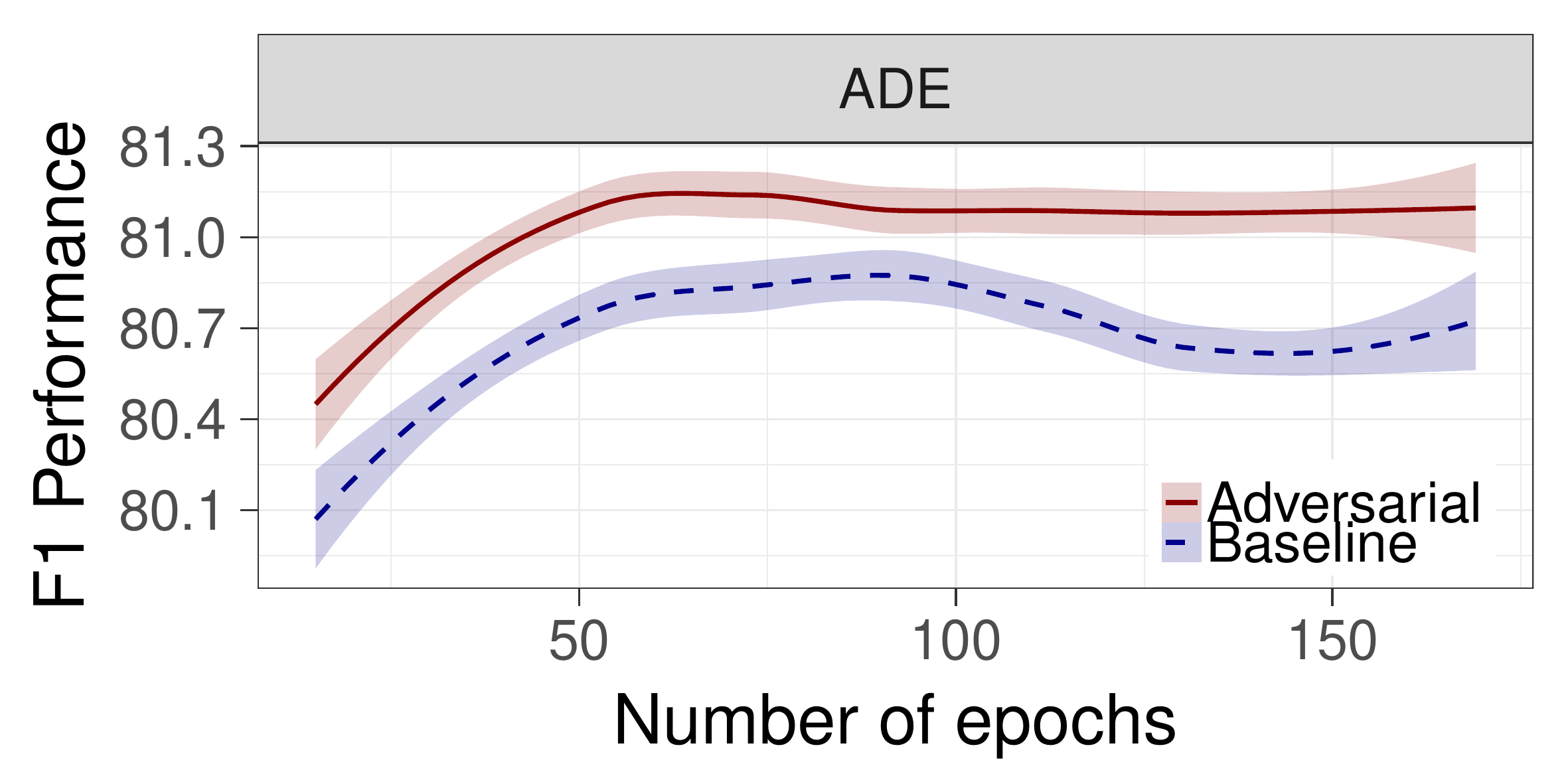}
  %\caption{ADE}
  %\label{fig:sub4}
        \end{subfigure}
\caption{F$_1$ performance of the baseline and the AT models on the validation sets from 10-30 epochs onwards depending on the dataset. The smoothed lines (obtained by LOWESS smoothing) model the trends and the 95\% confidence intervals.}
%\vspace{-2em}
\label{fig:at_vs_baseline}
\end{figure*}

\section{Results}
\label{sec:results}
\noindent \tabref{tab:results} shows our experimental results. The name of the dataset is presented in the first column while the models are listed in the second column. The proposed models are the following:
\begin{enumerate*}[label=(\roman*)]
\item \emph{baseline}: the baseline model shown in Fig.~\ref{fig:model} with the CRF layer and the sigmoid loss,
\item \emph{baseline EC}: the proposed model with the softmax layer for EC,
\item \emph{baseline (EC) + AT}: the baseline regularized using AT.
\end{enumerate*}
The final three columns present the F$_1$ results for the two subtasks and their average performance. 
Bold values indicate the best results among models that use only automatically extracted features. %) for each dataset are marked with bold font in \tabref{tab:results}.

For ACE04, the baseline outperforms \newcite{katiyar:17} by $\sim$2\% in both tasks. This improvement can be explained by the use of:
\begin{enumerate*}[label=(\roman*)]
\item multi-label head selection,
\item CRF-layer and
\item character level embeddings.
\end{enumerate*} 
Compared to \newcite{miwa:16}, who rely on NLP tools, the baseline performs within a reasonable margin (less than 1\%) on the joint task. On the other hand,~\newcite{li:17} use the same model for the ADE biomedical dataset, where we report a 2.5\% overall improvement. This indicates that NLP tools are not always accurate for various contexts.
For the CoNLL04 dataset, we use two evaluation settings. We use the \emph{relaxed} evaluation similar to~\newcite{gupta:16,heike:17} on the EC task. The baseline model outperforms % all  existing
the state-of-the-art models that do not rely on manually extracted features ($>$4\% improvement for both tasks), since we directly model the whole sentence, instead of just considering pairs of entities. Moreover, compared to the model of~\newcite{gupta:16} that relies on complex features, the baseline model performs within a margin of 1\% in terms of overall F$_1$ score.
We also report NER results on the same dataset and improve overall F$_1$ score with $\sim$1\% compared to~\newcite{miwa:14}, 
indicating that our automatically extracted features are more informative than the hand-crafted ones. 
These automatically extracted features exhibit their performance improvement mainly due to the shared LSTM layer that learns to automatically generate
feature representations of entities and their corresponding relations within a single model.
For the DREC dataset, we use two evaluation methods. In the \emph{boundaries} evaluation, the baseline has an improvement of $\sim$3\% on both tasks compared to~\newcite{bekoulis:18}, whose quadratic scoring layer complicates NER. 

\tabref{tab:results} and \figref{fig:at_vs_baseline} show the effectiveness of the adversarial training on top of the baseline model. In all of the experiments, AT improves the predictive performance of the baseline model in the joint setting. Moreover, as seen in~\figref{fig:at_vs_baseline}, the performance of the models using AT is closer to maximum even from the early training epochs. Specifically, for ACE04, there is an improvement in both tasks as well as in the overall F$_1$ performance (0.4\%). For CoNLL04, we note an improvement in the overall F$_1$ of 0.4\% for the EC and 0.8\% for the NER tasks, respectively. 
For the DREC dataset, in both settings, there is an overall improvement of $\sim$1\%. \Figref{fig:at_vs_baseline} shows that from the first epochs, the model obtains its maximum performance on the DREC validation set. Finally, for ADE, our AT model beats the baseline F$_1$ by 0.7\%.

Our results demonstrate that AT outperforms the neural baseline model consistently, 
considering our experiments across multiple and more diverse datasets than typical related works.
The improvement of AT over our baseline (depending on the dataset) ranges from $\sim$0.4\% to $\sim$0.9\% in
terms of overall F$_1$ score.
This seemingly small performance increase is mainly due to the limited performance benefit for the NER component, which is in accordance with the recent advances in NER using neural networks that report similarly small gains
(\eg the performance improvement
in \newcite{ma:16} and \newcite{lample:16} on the CoNLL-2003 test set is 0.01\% and 0.17\% F$_1$
percentage points, while in the work of \newcite{yasunaga:17}, a 0.07\% F$_1$ improvement on CoNLL-2000 using AT for NER is reported).
However, the relation extraction performance increases by $\sim$1\% F$_1$ scoring points, except for the
ACE04 dataset.
Further, as seen in \figref{fig:at_vs_baseline}, the improvement for CoNLL04 is particularly small on the evaluation set. This may indicate a correlation between the
dataset size and the benefit of adversarial training in the context of joint models, but this needs further
investigation in future work.

\section{Conclusion}
We proposed to use adversarial training (AT) for the joint task of entity recognition and relation extraction. The contribution of this study is twofold:
\begin{enumerate*}[label=(\roman*)]
\item investigation of the consistent effectiveness of AT as a regularization method over a multi-context baseline joint model, with 
\item a large scale experimental evaluation.
\end{enumerate*}
Experiments show that AT improves the results for each task separately, as well as the overall performance of the baseline joint model, while reaching high performance already during the first epochs of the training procedure. 

%\pagebreak

\section*{Acknowledgments}

We would like to thank the anonymous reviewers for the time and effort they spent in reviewing our work, and for
their valuable feedback.

\bibliographystyle{acl_natbib_nourl}
\bibliography{emnlp2018}

\end{document}